\def\mycmd{2}
\if\mycmd1
\documentclass[12pt,onecolumn, draftcls]{IEEEtran}
\else 
\documentclass[lettersize,journal]{IEEEtran} 
\fi
\usepackage{tabularx}
\usepackage{multicol}
\usepackage{multirow}
\usepackage{float}
\usepackage{mathtools}
\usepackage{amsmath}
\usepackage{amsthm}
\usepackage{amssymb}
\usepackage{graphicx}
\usepackage{transparent}
\usepackage{wasysym}
\usepackage{setspace}
\usepackage{color}
\usepackage{bm}
\usepackage{cases}
\usepackage{booktabs}
\usepackage{adjustbox}
\usepackage{subfigure}
\usepackage{glossaries}

\usepackage{tikz}

\usepackage{amsfonts}
\usepackage{makecell}
\usepackage{pbox}
\usepackage{epsfig}
\usepackage[bookmarks]{hyperref} 

\makeatletter

\floatstyle{ruled}
\newfloat{algorithm}{tbp}{loa}
\providecommand{\algorithmname}{Algorithm}
\floatname{algorithm}{\protect\algorithmname}

\theoremstyle{plain}

\theoremstyle{plain}

\theoremstyle{plain}

\usepackage{epsfig}
\usepackage{cite}
\usepackage{stfloats}
\usepackage{graphicx}
\usepackage{array}
\usepackage{enumerate}
\usepackage{enumitem}
\setlist[itemize]{leftmargin=*}
\setlist[enumerate]{leftmargin=*, label=\arabic*)}
\usepackage{graphicx}
\usepackage{times}
\usepackage{algorithm}
\usepackage{algpseudocode}
\theoremstyle{remark}

\makeatother

\newcolumntype{P}[1]{>{\centering\arraybackslash}p{#1}}
\newcolumntype{C}[1]{>{\centering\arraybackslash}m{#1}}

\algrenewcommand\algorithmicindent{1.0em}%
\providecommand{\lemmaname}{Lemma}
\providecommand{\propositionname}{Proposition}

\providecommand{\theoremname}{Theorem}
\providecommand{\theoremname}{Definition}
\newcommand{\rom}[1]{\uppercase\expandafter{\romannumeral #1\relax}}

\algnewcommand{\IIf}[1]{\State\algorithmicif\ #1\ \algorithmicthen}
\algnewcommand{\ElseIIf}[1]{\algorithmicelse\ #1} 
\algnewcommand{\EndIIf}{\unskip\ \algorithmicend\ \algorithmicif}

\newcounter{problem}
\newcounter{save@equation}
\newcounter{save@problem}


\numberwithin{save@problem}{subsection}
\numberwithin{save@equation}{subsection}

\begin{document}
\onecolumn
{\Large \textbf{Notice:} This work has been submitted to the IEEE for possible publication. Copyright may be transferred without notice, after which this version may no longer be accessible.}
\clearpage

\twocolumn
\title{Large Multimodal Models-Empowered Task-Oriented Autonomous Communications: Design Methodology and Implementation Challenges}
\author{Hyun Jong  Yang,~\IEEEmembership{Senior Member,~IEEE}, Hyunsoo Kim,~\IEEEmembership{Student Member,~IEEE}, Hyeonho Noh,~\IEEEmembership{Member,~IEEE}, Seungnyun Kim,~\IEEEmembership{Member,~IEEE}, and Byonghyo Shim,~\IEEEmembership{Fellow,~IEEE}
     \thanks{
     This work was supported, in part, by the National Research Foundation of Korea under Grant RS-2022-NR070834 and by the Institute of Information, Communications Technology Planning and Evaluation grant funded by the Korea government under Grant RS-2024-00398157 and Grant IITP-2025-RS-2024-00418784.
     H. Yang, H. Kim, and B. Shim are with Seoul National University, Korea (email: hjyang@snu.ac.kr, hskim@islab.snu.ac.kr, bshim@snu.ac.kr).
     H. Noh is with Hanbat National University, Korea (email: hhnoh@hanbat.ac.kr).
     S. Kim is with Massachusetts Institute of Technology, USA (email: snkim94@mit.edu).
    }
}

\maketitle
\begin{abstract}
Large language models (LLMs) and large multimodal models (LMMs) have achieved unprecedented breakthrough, showcasing remarkable capabilities in natural language understanding, generation, and complex reasoning.
This transformative potential has positioned them as key enablers for 6G autonomous communications among machines, vehicles, and humanoids.
In this article, we provide an overview of task-oriented autonomous communications with LLMs/LMMs, focusing on multimodal sensing integration, adaptive reconfiguration, and prompt/fine-tuning strategies for wireless tasks. We demonstrate the framework through three case studies: LMM-based traffic control, LLM-based robot scheduling, and LMM-based environment-aware channel estimation. From experimental results, we show that the proposed LLM/LMM-aided autonomous systems significantly outperform conventional and discriminative deep learning (DL) model-based techniques, maintaining robustness under dynamic objectives, varying input parameters, and heterogeneous multimodal conditions where conventional static optimization degrades.

\end{abstract}


\newacronym{USRA}{USRA}{user scheduling and resource allocation}
\newacronym{UL MU-MIMO}{UL MU-MIMO}{uplink multi-user multi-input multi-output}
\newglossaryentry{MU-MIMO}{
name=MU-MIMO,
description={a},
parent=UL MU-MIMO
}
\newglossaryentry{MIMO}{
name=MIMO,
description={a},
parent=UL MU-MIMO
}

\if\mycmd1 \newpage \else \fi

\section{Introduction}
\label{sec:1}
Driven by the huge success of ChatGPT, large language models (LLMs) have gained widespread attention, reshaping various fields by solving problems with zero-shot or few-shot prompting. Recently, large multimodal models (LMMs) extend this capability by embracing various modalities such as images, videos, and audio. These models can handle changing objectives and input variations using diverse multimodal observations. Domain-specific systems such as \emph{Talk2Drive} and \emph{RT-2} demonstrate the potential of LLM/LMM-powered control, reducing human takeover rates in autonomous driving and generalizing to unseen robot tasks.

To support autonomous services in practical applications such as logistics, manufacturing, healthcare, and agriculture, it is imperative to redesign entire end-to-end systems, including communication infrastructures. Needless to say, central to these autonomous tasks is the reliable and swift exchange of multimodal sensing data, intermediate states, and decision-making information. 
While significant efforts have been made to integrate LLMs/LMMs into the standalone autonomous agent such as humanoids or autonomous vehicles, their integration into wireless communication systems remains relatively unexplored. For simple and lightweight artificial intelligence (AI) tasks such as object detection or classification, AI functionalities can be directly embedded into autonomous agent. However, for more complex and resource-intensive AI tasks that LLM/LMM needs to handle, it is more reasonable to offload the computation to the external server. Future communications will transcend basic data transfer to include sophisticated functionalities such as multimodal data fusion, real-time adaptive reconfiguration, and holistic AI-driven decision-making. In such cases, it is natural to perform LLM/LMM processing at server, cloud, or data center. We henceforth call the place to perform the AI processing as central unit (CU).

Consider a smart factory where machines and robots communicate with the CU to execute tasks. After receiving multimodal sensing data and status information, the LLM/LMM in the CU generates on actions such as product scheduling, traffic control, and defect diagnosis, and then transmits them to the agents. After executing the tasks, autonomous agents send updated status and sensing data back to the CU. We call this paradigm \emph{task-oriented autonomous communications}, where infrastructures perform the given tasks by the help of autonomous AI and communication operations. This approach is expected to redefine wireless systems, especially in the 6G era, as autonomous machines become central to industrial applications.

An aim of this article is to present a comprehensive overview of LLM/LMM-powered wireless communication systems, focusing on design methodologies and practical implementation challenges. Table~\ref{tab:1} compares our approach with previous studies in~\cite{raiaan2024review,zhou2024large1,qu2024mobile1,boateng2024surveylargelanguagemodels,xu2024integration,javaid2024large,javaid2024leveraging}. We demonstrate the framework’s benefits through three representative case studies: i) an LMM-based vehicular network controller for dynamic traffic signal optimization, ii) a prompt-tuned LLM for robot scheduling in rapidly varying wireless environments, and iii) an environment-aware LMM-aided channel estimation using multimodal sensing and radio measurements. Our experimental results demonstrate notable gain of the task-oriented autonomous communication framework, including 79\% average speed improvement in vehicular networks, 47\% sum-rate increase in robot scheduling, and 4\,dB gain in channel estimation quality.

\begin{table*}[!t]
\centering
\caption{Comparison with Existing Surveys \& Magazines on LLM/LMM-aided Wireless Systems}
\label{tab:1}
\fontsize{10}{12}\selectfont
\begin{tabular}{|C{1.6cm}|C{7cm}|C{2.2cm}|C{2.4cm}|C{2.7cm}|}
\hline
Reference & Core Contribution & Various Modalities & Dynamic Task Adaptability & High-level System Requirements \\ 
\hline
\cite{raiaan2024review} 
& Comprehensive surveys on LLM architectures, capabilities, applications, and benchmarks 
& Not considered 
& Not considered 
& Not considered \\ 
\hline
\cite{zhou2024large1,qu2024mobile1,boateng2024surveylargelanguagemodels} 
& Application-specific surveys addressing telecommunications, mobile edge computing, and service management using LLMs 
& Text/limited sensing modalities	
& Static configuration	 
& Technical KPIs only \\ 
\hline
\cite{xu2024integration} 
& Exploration of LLMs/LMMs applications in transportation, autonomous driving, and Internet of Vehicles (IoV) 
& Vision/radar modalities
& Vision/radar modalities
& Not considered \\ 
\hline
\cite{javaid2024large,javaid2024leveraging} 
& Investigating LLM integration for unmanned aerial vehicle (UAV) and integrated satellite-aerial-terrestrial networks 
& Not considered 
& Predefined scenarios
& Not considered \\ 
\hline
\textbf{Our work} 
& \textbf{LLM/LMM-based task-oriented autonomous communications integrating multimodal sensing, adaptive task handling, and holistic orchestration} 
& \textbf{Considered} 
& \textbf{Context-aware reconfiguration	} 
& \textbf{Task completion metrics} \\ 
\hline
\end{tabular}
\vspace{-1em}
\end{table*}

The rest of this article is organized as follows: Section~\ref{sec:2} covers design methodologies, Section~\ref{sec:3} presents case studies and results, and Section~\ref{sec:4} outlines open issues and future directions.

\section{LLM/LMM-based Wireless Systems}
\label{sec:2}

\subsection{AI/ML-Aided Wireless Systems}\label{sec:2.A}
Keep pacing with the advancement of AI, the wireless industry has begun integrating AI into standardization. Since Release 18~\cite{3gppTR38843}, 3GPP has studied AI/machine learning (ML) functions in the radio interface, access network, and core network. The RAN1 group identified three use cases—beam management, CSI feedback, and positioning—for standardization in Release 19. 3GPP also outlined AI/ML management in 5G systems, covering model training, validation, deployment, and retraining~\cite{3GPP_TR_28.908}. While the considered AI/ML techniques work for predefined scenarios~\cite{Hyoungju18, Haque2023}, they lag behind rapid AI advances and are insufficient for task-oriented autonomous communications. As autonomous agents are mobile, have diverse goals, and face abrupt task changes, more elaborated AI techniques are needed for next-generation wireless systems.

\begin{figure*}
    \centering    \includegraphics[width=\if\mycmd1 0.78 \else 0.78\fi\textwidth]{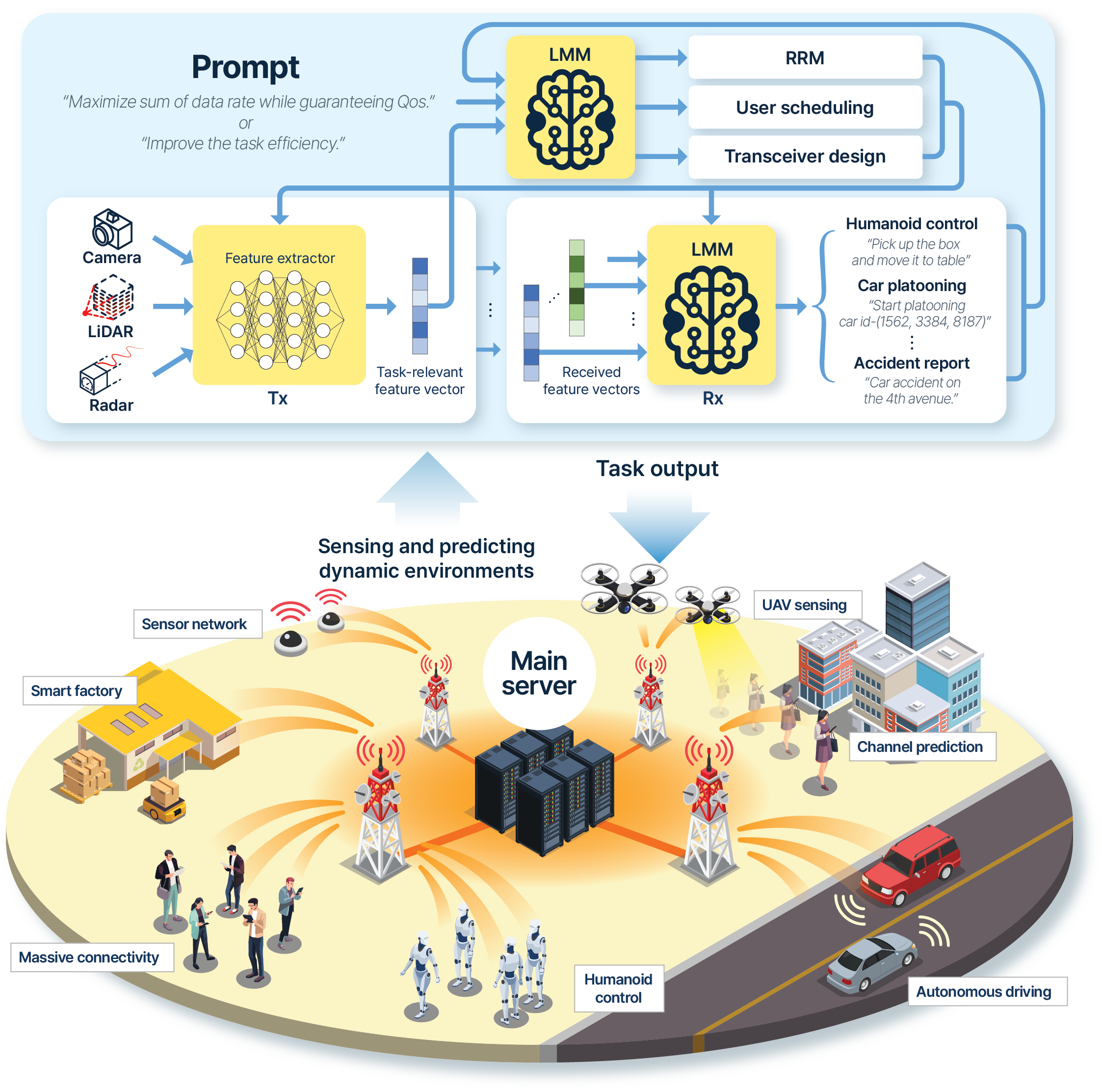}
    \caption{LLM/LMM-based system design for task-oriented communications.}
    \label{fig:1}
    \vspace{-1em}
\end{figure*}

\subsection{LLM/LMM-Based Task-Oriented Communication Systems: Goals and Characteristics} \label{sec:2.B}
In task-oriented communication systems (Fig.~\ref{fig:1}), robots, vehicles, and machines send multimodal sensing data to the CU in compressed form (e.g., quantized feature vectors or text tokens). Internal statuses and intermediate results (e.g., position, direction, packet error rate, channel state information, battery level) are also fed back, enabling LLMs/LMMs to update subtask goals based on the aggregated information.

The salient features of LLM/LMM-based autonomous communication systems are as follows. First, to collect, deliver, and interpret multimodal sensing and control data and then control the autonomous agents, communication systems need to be redesigned. CU inputs come from diverse sensing modalities (e.g., camera, LiDAR, radar, inertial measurement unit (IMU), radio frequency), providing rich environmental context, while outputs to agents may be binary control signals, text, or even voice.

Secondly, key performance indicators (KPIs) must be redefined to check the completion of given task. One can easily notice that conventional metrics like bit error rate (BER) or spectral efficiency cannot capture the operational success of task-oriented communications. So, to reflect more abstract and task-related requirements, such as ``control traffic signs in an intersection to maximize the average vehicle speed'', the task completion accuracy should be accounted for in the KPI design.

Thirdly, objectives for multiple autonomous agents must adapt dynamically to environmental changes and agent status. For example, surgery robots should respond immediately to a sudden drop of the patient’s blood pressure, while car assembly lines may reprioritize tasks when parts are in short supply. Such changes require rapid, adaptive reconfiguration of communication and network parameters while preserving overall task goals.

While LLM/LMMs offer a number of benefits, their deployment in wireless systems is constrained by several technical hurdles. Delivering high-volume multimodal data streams must meet stringent bandwidth, latency, and reliability constraints. Sensing pipelines must remain robust and synchronized despite occlusions, noise, and environmental shifts. To sustain real-time operation, LLM/LMM computation needs to be properly allocated to either edge or cloud. Recognizing these challenges and constraints is the first step for designing practical LLM/LMM-based task-oriented communication systems.

\begin{figure*}[tbh!]
\centering    \includegraphics[width=\if\mycmd1 0.84 \else 0.85\fi\textwidth]{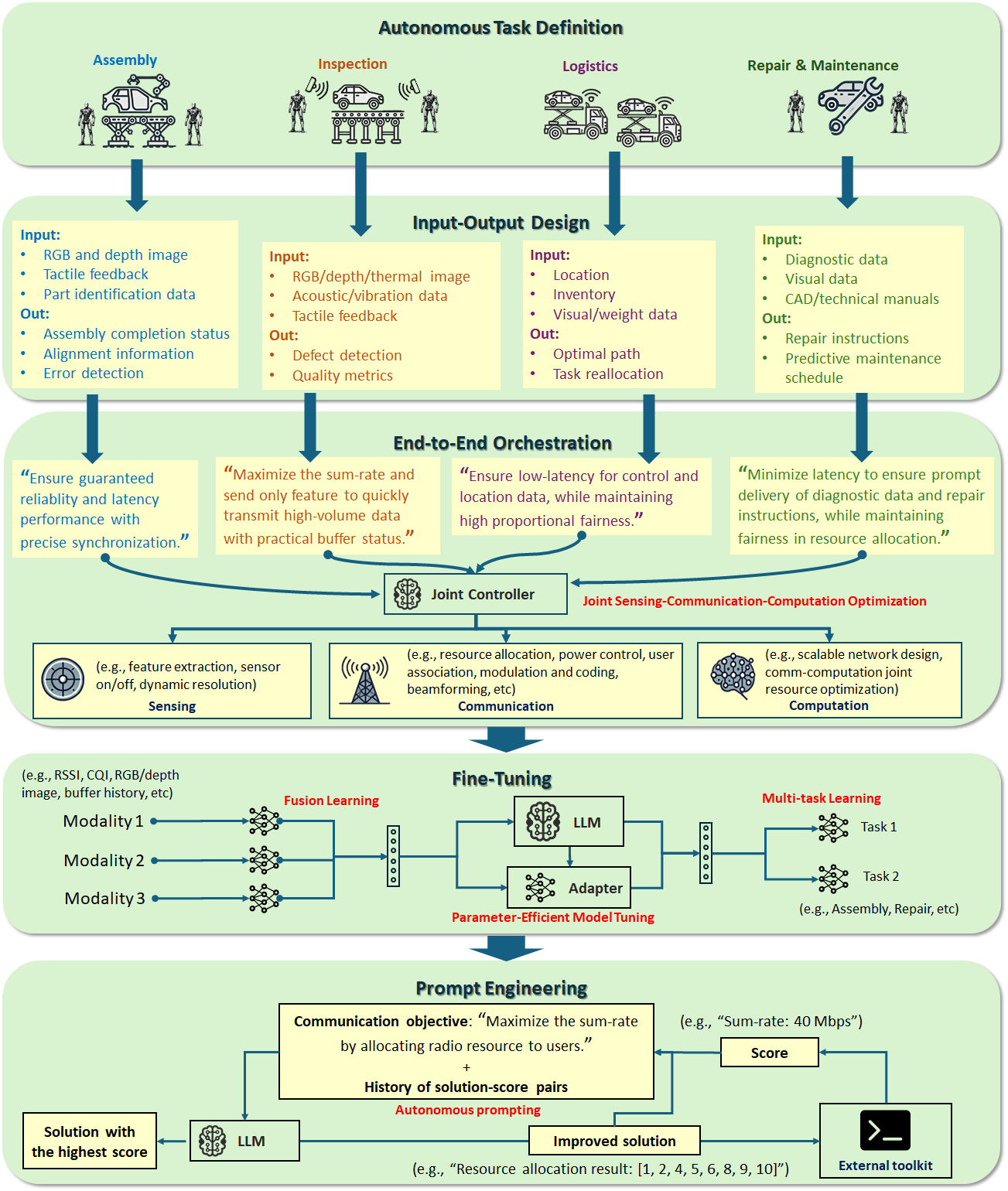}
\caption{Design procedure of LLM/LMM-based task-oriented communications.}
\label{fig:2}
\vspace{-1em}
\end{figure*}
    
\subsection{LLM/LMM-Based Wireless Systems: Design Methodology} \label{sec:2.C}
The key steps to design LLM/LMM-based task-oriented communication systems are illustrated in Fig.~\ref{fig:2}, with each stage briefly explained below:

\begin{itemize}
    \item \textbf{Task design:} Since LLM/LMM-based tasks are expressed in a sentence (e.g., ``guarantee quality of service (QoS) for each agent while maximizing proportional fairness (PF)''), it would be natural to use the sentence-based task generator—potentially implemented by a language model. For example, in a disaster, tasks might be ``deploy drones to survey areas, identify victims, and prioritize rescue by severity.'' Note that the set of tasks assigned by the CU does not have to be predefined or finite since the CU interprets the received multimodal sensing data to infer the environmental context and then dynamically formulates or adjusts task objectives. For example, in robot scheduling, the CU may generate a scheduling task that prioritizes proportional fairness when uplink buffers are congested but switches to sum-rate maximization when the minimum QoS requirement is satisfied.

    \item \textbf{Input-output design:} A key feature of task-oriented autonomous communications is integrating AI operations with wireless links. Because LLM/LMM inputs include multimodal sensing data, these must be compressed (e.g., via embedding and quantization) before the uplink transmission. Generating features that exclude irrelevant background improves task completion, and sensing data may even be converted into text tokens. Unlike conventional cellular systems, uplink traffic dominates while downlink mostly carries simple commands. While these steps outline the functional flow, each stage poses distinct technical challenges. High-volume multimodal data must be compressed and transmitted without degrading task-critical information. Sensing streams require robust feature extraction and synchronization to avoid performance drops from noise, occlusions, or asynchronous updates. On the computing side, the orchestration logic must decide when to process locally or offload to the central unit, to optimize the latency, energy, and model accuracy.

    \item \textbf{End-to-end system orchestration:} Conventional systems optimize each block separately using the rule-based scheduling with feedback (e.g., CQI, RI, PMI in LTE, or SSB in 5G). In contrast, LLMs/LMMs determine system parameters holistically for all agents, requiring training on diverse datasets that map inputs, states, and task objectives to complete system configurations. For example, in a smart factory, an LMM orchestrates end-to-end AGV communication by prioritizing the uplink transmissions from critical areas based on sensor and network data and offloading urgent network tasks to edge servers, which can be implemented via the user scheduling, resource allocation, and beamforming.

    \item \textbf{Fine-tuning:} Fine-tuning adapts pre-trained models (e.g., LLaMA, Gemini, LLaVA) to task-specific datasets, enabling learning of domain knowledge absent in pre-training. Commonly used approaches include few-shot, transfer, and adapter learning. In low-rank adaptation (LoRA), for example, model parameters remain frozen while only a small low-rank adapter block is trained (see Section~\ref{sec:3.A}).

\item \textbf{Prompt engineering:} Retraining all LLM/LMM parameters whenever task or environment changes is by no means possible. Prompt engineering enables rapid adaptation without weight updates by feeding a few tailored prompts. A popular approach is chain-of-thought (CoT) prompting, which breaks complex tasks (e.g., resource allocation under QoS constraints) into smaller steps, allowing reasoning checkpoints where corrective prompts can be injected. For example, if QoS or bandwidth limits are violated, CoT can be guided to ``set unsatisfied user's rate to zero and optimize PF for the rest."
\end{itemize}

\begin{figure*}
\centering    \includegraphics[width=\if\mycmd1 0.89 \else 0.85\fi\textwidth]{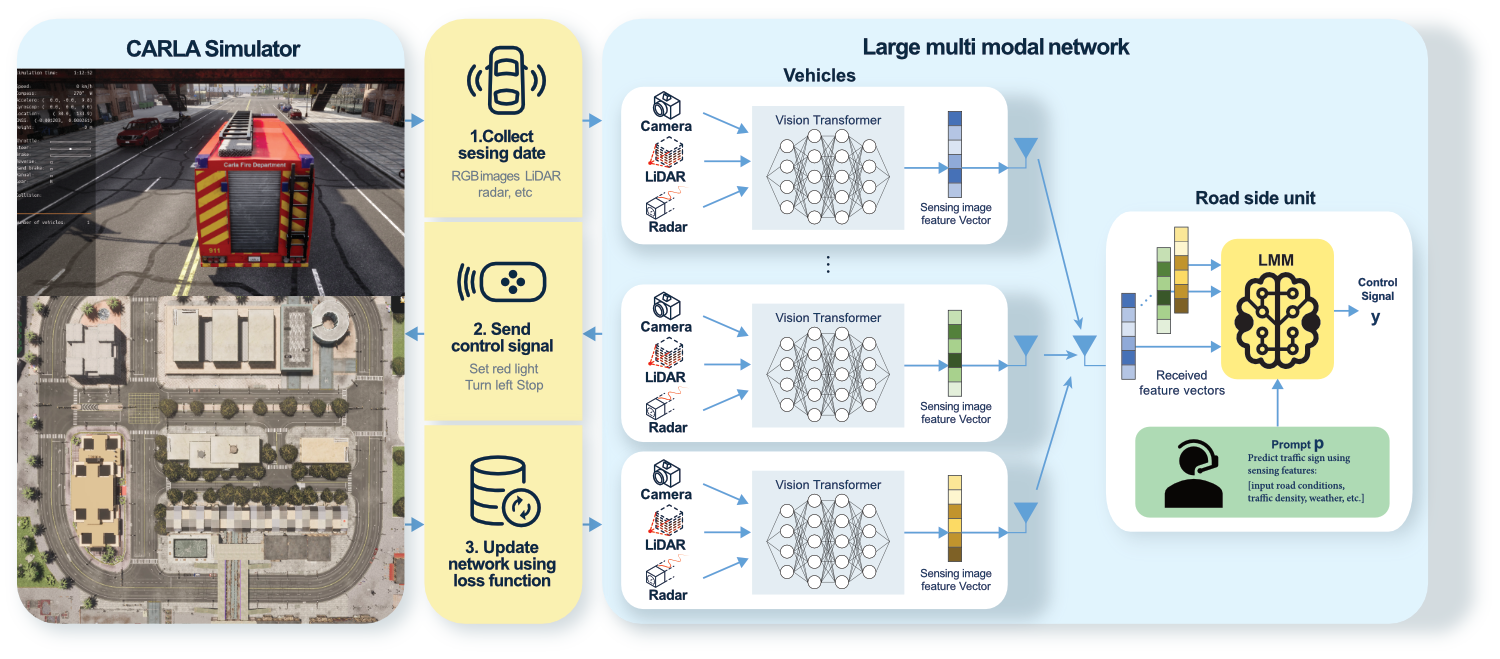}
\caption{Illustration of traffic sign control using fine-tuned LMM.}
\label{fig:3}
\vspace{-1em}
\end{figure*}

\section{Case Studies and Performance Evaluation} \label{sec:3}
To validate the proposed approach, we design three CU-controlled autonomous systems: (i) an LLM-aided traffic control for vehicle-to-everything (V2X) networks, (ii) an LMM-assisted channel prediction using vision sensing, and (iii) an LLM-based user scheduling in dynamically varying wireless environments.

\subsection{LMM-Aided V2X Communications for Vehicle Traffic Control} \label{sec:3.A}
An aim of the LMM-based traffic sign control task is to minimize the vehicle latency at intersections (Fig.~\ref{fig:3}). In the proposed \textit{LMM-based V2X communications} (LMM-V2X), each vehicular user equipment (VUE) uses a Vision Transformer (ViT)~\cite{vit} to extract key traffic features from sensing images, then sends quantized vectors to the roadside unit (RSU), where the LMM generates speed-maximizing traffic signals. To reduce the signaling overhead from high-fidelity sensing data, an encoder prioritizing foreground features (e.g., vehicle and pedestrian) is employed. In doing so, foreground data can be transmitted more frequently, reducing feedback load without compromising decision quality.

\noindent \textbf{LoRA fine-tuning:} To generate optimal traffic control signals, we perform the fine-tuning of the pre-trained TinyLLaVA model (3.1B parameters)~\cite{zhou2024tinyllava1}. We use LoRA which updates only small low-rank matrices while keeping the main weights frozen. While significantly relieving the fine-tuning overhead, LoRA also helps identify which parts of the model drive traffic control changes, aiding adaptation to unseen road conditions.

\noindent \textbf{CoT prompting:} By directly incorporating real-time information on road conditions, network states, and environmental changes into the CoT prompt, the LMM can not only adjust its own decision-making but also coordinate sensing, communication, and computation resources across all VUEs.

\noindent \textbf{V2X Dataset generation:} Dataset is generated using the CARLA simulator, a cyber-physical platform for autonomous driving~\cite{Dosovitskiy2017}. The simulated V2X scenario places the RSU at an intersection corner in a $100\,\text{m} \times 100\,\text{m}$ area with $20\,\text{m}$ lanes. $20$ $\sim$ $50$ VUEs are randomly positioned in four approach directions, each selecting a traffic signal action (left turn, straight, or right turn). For fair comparison, the LMM-V2X and CNN baseline use the same set of VUE-view images (identical capture settings and timing). LMM-V2X uses quantized features derived from these images, whereas the CNN at the RSU is based on the raw images; no extra metadata or additional views are provided, so both pipelines operate on the same amount and type of input, ensuring apple-to-apple comparison. By contrast, LMM-RSUview uses only RSU top-view images without VUE feedback.
The simulation environment assumes a crash-free scenario, where traffic dynamics exclude any car crashes or other safety-critical events.
For training and evaluating LMM-V2X, we generate the dataset using the NR V2X channel model in 3GPP TR 38.885. Simulations run on a system with an Intel Xeon Gold 6326 CPU (16 cores) and NVIDIA L40S GPUs (48\,GB VRAM).

\noindent \textbf{Benchmark schemes:} We compare the average vehicle speed performance of the proposed LMM-V2X with three baseline schemes: 1) convolutional neural network (CNN)-based technique, which uses CNN at RSU for the traffic control, 2) the LMM-RSUview controller where the LMM in RSU uses the captured top-view images of the road in the intersection to control the traffic lights, and 3) the conventional round-robin scheme that repeats the fixed traffic light sequence with equal time intervals. 

\begin{figure}[!t] 
    \centering 
    \includegraphics[width=0.9 \columnwidth]{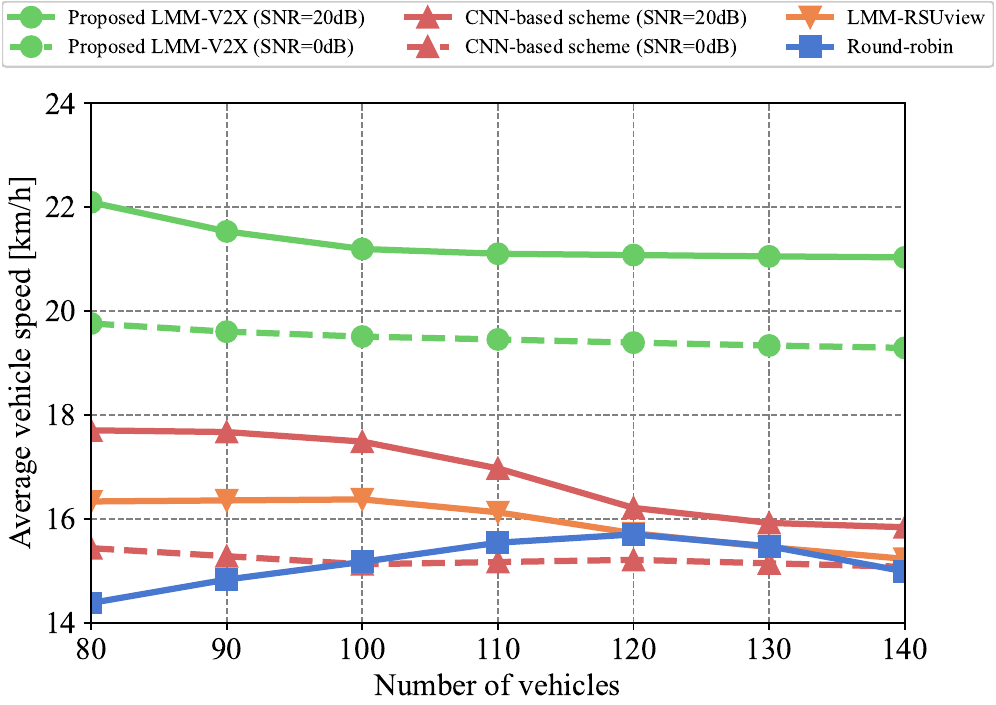}  \caption{Average vehicle speed as a function of the number of vehicles.} 
    \vspace{-1em}
    \label{fig:4}
\end{figure}

\noindent \textbf{Performance evaluation:} 
Fig.~\ref{fig:4} shows the average VUE speed versus the number of vehicles. LMM-V2X outperforms conventional schemes by a large margin. For example, with 80 vehicles, LMM-V2X achieves over 79\% speed improvement over the round-robin scheme. Even when compared to LMM-RSUview, LMM-V2X improves the speed by over 30\%, as LMM-RSUview depends solely on a single RSU-mounted camera (typically several meters above the ground), which can be blocked by tall vehicles or miss details in corners or dead zones. In contrast, due to multi-view images obtained from all VUEs, LMM-V2X can capture the full intersection context and issue more reliable control commands. Under identical VUE-view inputs and a fixed number of vehicles, the gain mainly comes from decision composition since the controller fuses per-vehicle front-view observations into a coherent global state and plans discrete signal phases.

\begin{figure*}[!t] 
    \centering 
    \includegraphics[width=\if\mycmd1 1 \textwidth \else 0.9\textwidth \fi]{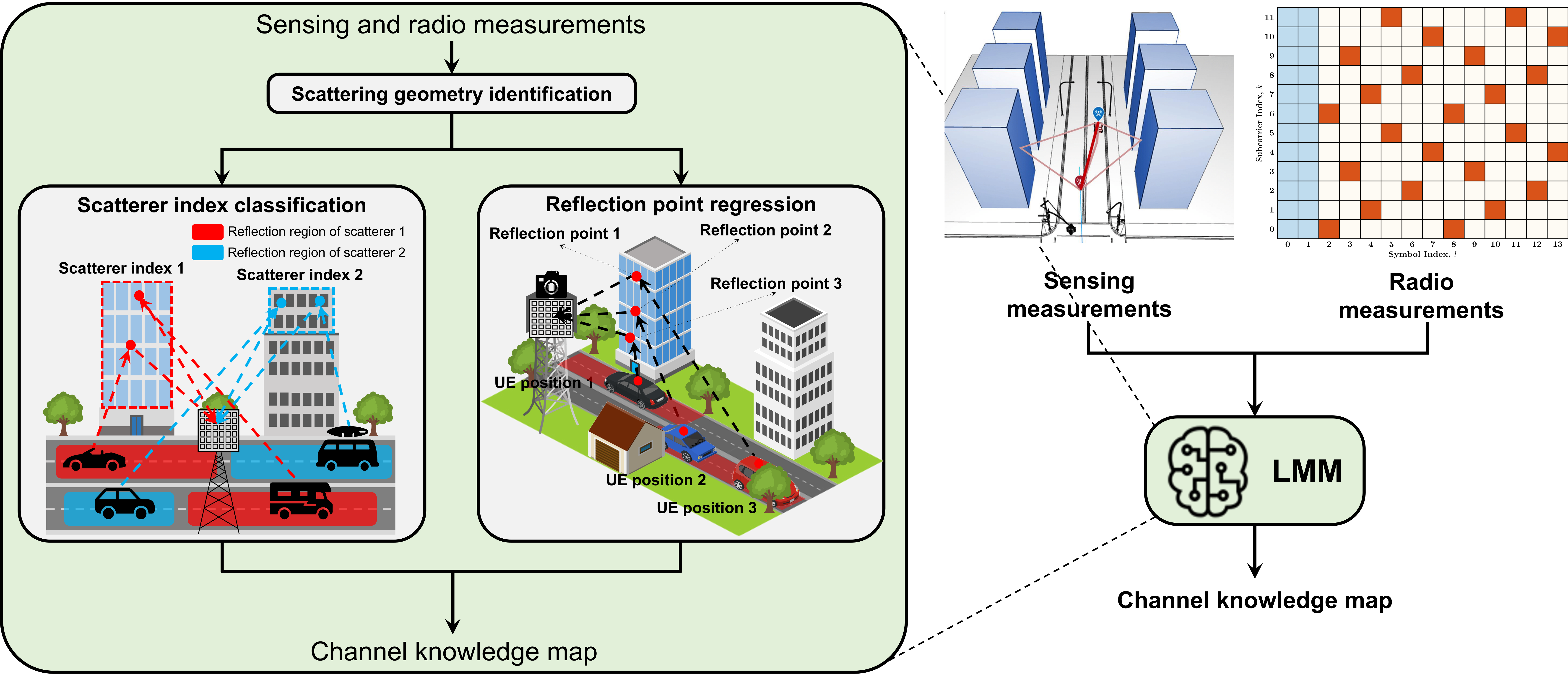} \caption{Illustration of LMM-based environment-aware channel estimation.} 
    \vspace{-1em}
    \label{fig:5}
\end{figure*}

\noindent \textbf{Implementation challenges:} In a real V2X deployment, transmitting multi-view images or extracted features from numerous VUEs to the RSU would be subject to fluctuating channel quality, interference, and bandwidth limits. Feature extraction must remain robust despite occlusions from tall vehicles and variable viewpoints since any misalignment can degrade the LMM’s control accuracy.

\subsection{LMM-Based Environment-Aware Channel Estimation} \label{sec:3.B}
In this study, we explore an LMM-based environment-aware channel prediction that integrates multimodal sensing and radio measurements to model the surrounding environment and then predict signal propagation paths. The \textit{LMM-based environment-aware channel estimation} (LMM-CE) learns the intrinsic mapping, termed the channel knowledge map (CKM), between user position and CSI. Since CSI has high dimensionality that scales with the number of antennas, direct learning of the full end-to-end mapping is computationally infeasible. To address this, we introduce an intermediate scattering geometry—estimating the locations of scatterers and reflection points—as a bridge between user position and CSI. LMM-CE operates in two stages: 1) classifying scatterers that contribute to signal reflection for a given user position, and 2) regressing the corresponding reflection points on those scatterers (see Fig.~\ref{fig:5}). By unifying sensing interpretation, environmental modeling, and channel state inference into a single decision-making engine, the LMM serves as an end-to-end system orchestrator coordinating uplink feature compression, environment-aware CSI prediction, and downlink configuration support. In our simulations, we employ a ray tracing-based urban channel model and utilize a system equipped with Intel Xeon Gold 6326 CPU (16 cores) and NVIDIA L40S GPUs (48\,GB VRAM).

\noindent \textbf{LMM prompting and fine-tuning}  
To learn the scattering geometry, we design stage-specific prompts for LMM-CE: 1) \emph{scatterer index classification}, where each prompt provides the user’s position, a semantic map of the environment, and candidate scatterer locations, instructing the LMM to identify which objects act as scatterers (e.g., “Perform a binary classification to determine whether the object is a scatterer for a given user position. An object is a scatterer if there is a reflection point satisfying the law of reflection with respect to the base station (BS) and user”), and 2) \emph{reflection point regression}, where prompts include the user’s position, scatterer location, and a known reflection point, guiding the LMM to predict the point on the surface that satisfies the law of reflection (e.g., “Perform a regression task to determine the reflection point given the user position. The point must satisfy the law of reflection with respect to the BS and user”). We employ the LLaVA-NeXT-Interleave 7B model and perform the LoRA-based fine-tuning.

\noindent \textbf{Dataset generation:}  
We use MATLAB R2024B to create sensing and radio measurement datasets. First, we model 3D urban environments with up to four buildings along a four-lane road parallel to the $x$-axis using MATLAB's triangulation functions (see Fig.~\ref{fig:5}). Each building and lane is $6\,\text{m}$ and $2\,\text{m}$ wide, respectively. Next, the \texttt{raytrace} function computes signal propagation paths and extracts multipath parameters (angles, delays, path gains) for pilot data. Finally, visual sensing data is generated with MATLAB's \texttt{siteviewer}, using $1024 \times 768$ resolution and a $120^\circ$ field of view.

\noindent \textbf{Benchmark schemes:}
We compare the performance of the proposed LMM-CE with five competing schemes: 1) LMM-based geometric channel parameter (GCP) regression technique that estimates GCPs such as angles, delays, and path gains from user positions, 2) ViT-based VCP regression technique, 3) CNN-based VCP regression scheme, 4) Transformer-based GCP regression technique, and 5) linear GCP regression technique based on linear regression method.

\begin{figure}[!t] 
    \centering 
    \includegraphics[width= 1\columnwidth]{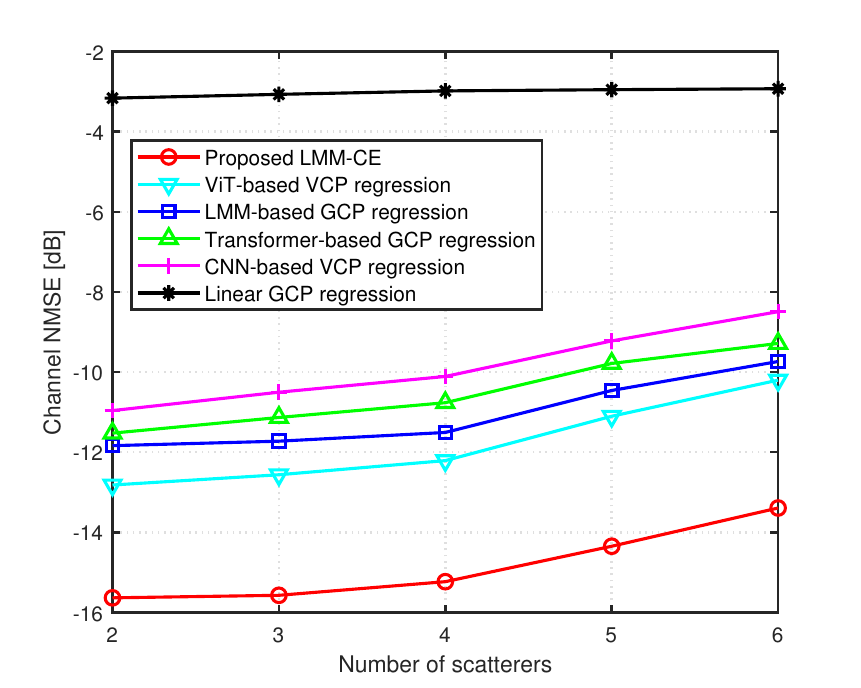} \caption{\textbf{Channel NMSE as a function of the number of scatterers.}}
    \vspace{-1em}
    \label{fig:6}
\end{figure}

\noindent \textbf{Performance evaluation:}
In Fig.~\ref{fig:6}, we evaluate the channel prediction performance in terms of normalized mean squared error (NMSE). We observe that the proposed LMM-CE outperforms all baseline schemes in both scenarios. For example, when the number of scatterer is $4$, LMM-CE achieves a $4\,$dB gain over the Transformer-based GCP regression technique. In LMM-CE, the LMM parses the natural language instructions to comprehend the underlying physical law (i.e., the law of reflection) between user position and scattering geometry so that it achieves substantial gain in channel prediction accuracy.

\noindent\textbf{Implementation challenges:} For live environment-aware channel estimation, the joint use of visual sensing and radio measurements would face challenges from incomplete or noisy sensing data, as well as rapid changes in scatterer positions or surface properties. The ray-tracing-based CKM generation assumes static geometry, but in practice moving objects and environmental changes might lead to mismatches between predicted and actual channels. Running the two-stage LMM inference for scatterer classification and reflection point regression for multiple users would also require optimized processing pipelines to avoid feedback and configuration delays.

\subsection{Dynamically-Varying Robot Scheduling using Fine-Tuned LLM} \label{sec:3.C}
In this study, we test the quality of LLM-based communication system in handling dynamically varying task objective of multi-robot operations. As a system orchestrator, CU continuously processes multimodal sensing information, such as camera images and LiDAR scans, to detect dynamic changes in the communication environment. Upon detecting such changes, CU formulates updated task goals and requirements as a natural-language prompt and supplies it to LLM for the real-time rescheduling. In this study, we assess the capability of the LLM in solving the scheduling optimization problem under such dynamically changing communication environments. Specifically, we measure the sum-rate and PF performances of the \textit{LLM-based robot scheduling} (\textit{LLM-RS}) where the target task switches between two distinct scheduling scenarios (henceforth referred to as Task 1 and Task 2 for brevity). Note that such task switching frequently occurs in autonomous operations in smart factories, farms, and warehouses.

Specifically, Task~1 targets fair uplink transmission of sensing data from robots to the CU. To this end, the LLM in the CU allocates resource blocks (RBs) to maximize proportional fairness (PF). Since the uplink data generation is sporadic, each robot’s buffer status is set randomly. Task~2 involves the CU sending action commands to each robot via downlink based on collected sensing data. Here, the primary objective is to satisfy each robot’s QoS requirement; once the minimum rate is met, remaining RBs are allocated to maximize the sum-rate of robot agents.

\noindent \textbf{Dataset generation:} In our simulations, we consider network environments where 10 to 50 robots are uniformly distributed within a radius of either 50 meters or 100 meters from the CU. Following the IEEE 802.11ax standard, we set the bandwidth to 20 MHz (corresponding to 9 RBs). The simulation is performed on a computer equipped with AMD Ryzen 5950X CPU (16 cores) and NVIDIA GeForce RTX 3090 (24 GB VRAM).

\begin{figure}[!t] 
    \centering
    \includegraphics[width= 1\columnwidth]{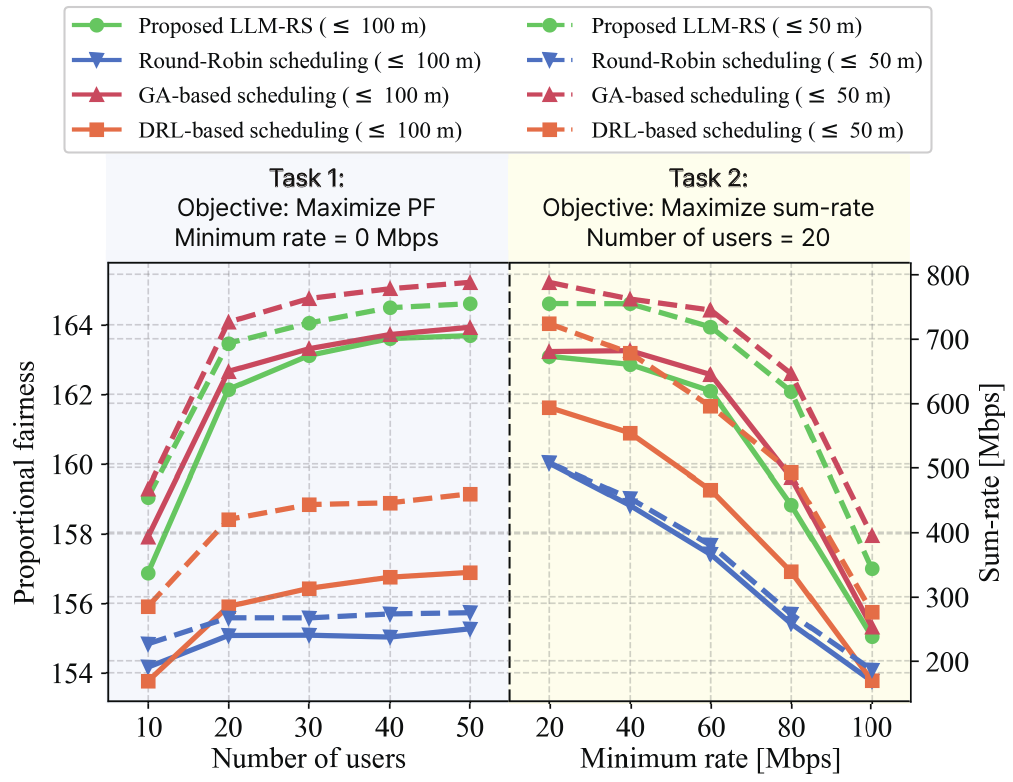} 
    \caption{\textbf{PF and sum-rate of LLM-RS and baseline methods in the dynamic environment where task, number of users, and minimum rate requirements continuously change. The same pre-trained LLM-RS is used without any retraining, even as the environment changes.}}
    \label{fig:7}
    \vspace{-1em}
\end{figure}

\noindent \textbf{LLM fine-tuning:} 
As a language-based model, the LLM cannot directly solve our scheduler optimization problem. We address this using \emph{optimization by prompting} (OPRO)~\cite{Yang20231}, which iteratively refines prompts to guide the LLM toward optimal solutions. In each iteration, feedback on the generated RB allocation is used to adjust the prompt, helping the model internalize the scheduling objective. For example, if LLM-RS outputs ``[1 2 4 5 6 8 9 10]'' and robots~3 and~7 fail to meet QoS, OPRO issues the prompt ``RB allocation vector violates the QoS requirement of robots 3 and 7,'' forcing regeneration. This process continues until all QoS constraints are satisfied. Even when the task objective is changed, LMM-RS can handle it by simply updating the prompt. Note that, due to the hallucination effect, the LLM may occasionally schedule robots that are not associated with the task or assign excessive RBs to a robot. To mitigate the problem, OPRO continuously checks whether the generated solution violates the scheduling description, and forces the LLM-RS to regenerate the RB allocation vector unless the solution conforms to the specified constraints.

Note that OPRO adopts a strategy analogous to simulated annealing: LLM-RS first explores a diverse range of solutions and incorporates the corresponding performance feedback from OPRO into the prompt, from which LLM-RS learns the underlying structure of the scheduling task. As the process advances, the LLM gradually shifts its focus toward generating higher-quality solutions based on the updated prompt. This balance between exploration and exploitation helps OPRO to escape from local optima and converge to high-quality feasible solutions.

\noindent \textbf{Benchmark schemes:} 
We test three benchmarks:  
1) \emph{Round-robin} scheduling, which assigns RBs equally to all users, 2) \emph{Deep reinforcement learning} (DRL)-based scheduling, where RB allocation is determined from the robots’ signal-to-noise-ratio (SNR) map with sum-rate or PF as the reward, and 3) \emph{Genetic algorithm} (GA)-based scheduling, which mimics natural selection through selection, crossover, and mutation to produce near-optimal solutions. While GA provides the best-achievable upper bound for our task, obtaining its solution requires exhaustive search and can take days or even months. All scheduling methods receive identical SNR information as input to ensure a fair comparison.

\noindent \textbf{Performance evaluation:} Fig. \ref{fig:7} illustrates the performance of LLM-RS and baseline schemes in dynamic environments where the target task switches from Task 1 to Task 2. We evaluate PF under the Task 1 scenario and sum-rate under the Task 2 scenario. Interestingly, LLM-RS performs close to GA for both tasks, achieving near-optimal scheduling performance. Considering that pre-trained LLMs are not optimized for solving mathematical optimization problems, it is a bit surprising to see the reasoning and inference power of a properly fine-tuned LLM. Since the DRL-based technique needs to explore the entire scheduling decision space from scratch, it often gets trapped in local optima, causing severe performance degradation.

As the target task switches from Task 1 to Task 2, the objective function and constraint are also changed. Under the newly imposed minimum QoS requirement, any robot whose achieved data rate falls below the prescribed threshold is treated as having zero data rate in the sum-rate computation. Even with the pre-trained LLM-RS in Task 1, LLM-RS significantly outperforms both DRL-based and round-robin schemes. When Task 1 is switched to Task 2, LLM-RS adjusts its RB allocation such that the sum-rate is maximized while ensuring the minimum QoS. Note that to accommodate such changes in the objective function and constraints, the GA-based technique should be re-optimized.

\noindent \textbf{Implementation challenges:} In operational multi-robot networks, uplink contention and variable link quality could delay status reports to the CU, deteriorating the LLM’s scheduling decisions. Task switches between the proportional fairness and sum-rate maximization may occur more frequently and under less predictable conditions, hindering the convergence of OPRO.

\section{Discussion and Challenges}
\label{sec:4}
In this article, we presented an LLM/LMM-based autonomous communication paradigm where tasks are autonomously executed without human intervention through the integration of generative AI, multimodal sensing, and wireless communications. We show that LLM/LMM serves as a holistic decision-maker and system optimizer for the given wireless tasks. we also demonstrate that the LLM/LMM-based controller exposes goals and constraints via prompts and leverages lightweight adapters (e.g., LoRA), enabling reuse across scenarios with minimal structural changes and markedly lower engineering/training overhead. While this study highlights key aspects and case studies, there are many open challenges requiring further investigation. We outline several future research directions:

\begin{itemize}
    \item \textbf{Integration with wireless communication standards:} Task-oriented communications influence all aspects of wireless systems, including functions, procedures, and interfaces in the 3GPP standardization process. A key challenge is that the training dataset accurately captures wireless environments and operational scenarios such as mobility, signaling, and QoS should align with 3GPP protocols. The dataset must be approved by standards to ensure a common baseline, although vendor-specific fine-tuning might be preferable to meet specific hardware and deployment needs.

    \item \textbf{Ultra-low latency inference and adaptive LLMs/LMMs:} Low-latency inference is crucial for task-oriented autonomous communications but ensuring timely execution poses practical challenges due to the large model size and memory footprint of LLMs/LMMs. Model compression techniques such as pruning, quantization, and knowledge distillation can reduce model complexity and computational load. Edge computing and task offloading, where simpler tasks such as beamforming are handled locally by smaller models and more complex tasks (e.g., user association) are processed by larger models at CU, will also be useful.

    \item \textbf{Addressing the hallucination problem:} Generative models like LLMs/LMMs can occasionally produce unrealistic or erroneous outputs, a phenomenon called hallucination, which can undermine wireless system reliability considerably. Mitigation requires balanced, high-quality datasets tailored for wireless contexts, verification mechanisms to ensure output accuracy, and low-complexity retraining with updated domain-specific knowledge (e.g., via imitation or safe learning). Also, a hybrid system architecture, integrating LLM/LMM-based systems and conventional rule-based systems, can be useful. Specifically, a validation module assesses the plausibility and effectiveness of LLM/LMM-generated outputs. If an output fails to meet predefined reliability criteria, a fallback mechanism is triggered to either re-run the LLM/LMM inference with adjusted prompts or revert to the conventional rule-based system.

    \item \textbf{Developing wireless foundation models:} Most LLM/LMM models are trained for general-purpose applications such as chatbots. To maximize domain-specific expertise and operational efficiency, it is desirable to develop wireless foundation models that are specifically designed, pre-trained, and fine-tuned for wireless tasks. Unlike general-purpose models trained primarily on text and images, wireless foundation models are purpose-built using wireless data (e.g., I/Q samples and CSI), equipping them with an intrinsic understanding of communication systems. This results in faster inference, lower computational overhead, and reliable decision-making in complex wireless environments.
    \item \textbf{Computation efficiency and power consumption:} While current LLMs/LMMs are general-purpose and often consume more computational resources and power than task-specific models, their flexibility in handling unseen data, integrating multiple modalities, and enabling rapid system design without bespoke optimization offers unique benefits for complex and dynamic wireless scenarios. From a hardware perspective, continuous advancement of NPUs and low-power accelerators will improve both throughput and energy efficiency. On the algorithmic side, emerging model compression and architecture optimization such as mixture of experts (MoE) techniques will further reduce the computational load and power usage.

\item \textbf{Security and adversarial robustness:} Since LLM/LMM-based systems can interact in natural language and accept multimodal inputs, they may be vulnerable to adversarial prompts or crafted inputs from attacker. This risk calls for integrated security measures, such as input sanitization and filtering, adversarial example detection, and role-based access control over critical system functions. Additional research on prompt-injection prevention, robust fine-tuning, and continuous monitoring of model outputs is essential to mitigate such vulnerabilities. In safety-critical wireless applications, LLM/LMM components should operate within a secured execution environment with strong authentication and encrypted communication links, and should be paired with fail-safe fallback mechanisms.

\end{itemize}


\begin{IEEEbiography}[{\includegraphics[draft=false,width=1in,height=1.5in,clip,keepaspectratio]{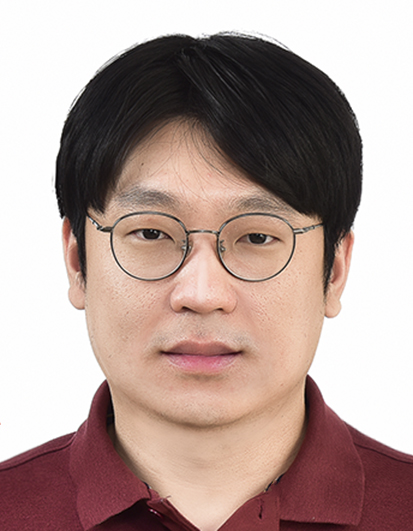}}]{\textit{Hyun Jong Yang}} (hjyang@snu.ac.kr) is an Associate Professor with the Department of Electrical and Computer Engineering, Seoul National University (SNU), Seoul, South Korea. He received the Ph.D. degree in electrical engineering from Korea Advanced Institute of Science and Technology, Daejeon, South Korea, in 2010. His fields of interests are signal processing, wireless communications, and machine learning. He is a senior member of IEEE.
\end{IEEEbiography}

\begin{IEEEbiography}[{\includegraphics[draft=false,width=1in,height=1.5in,clip,keepaspectratio]{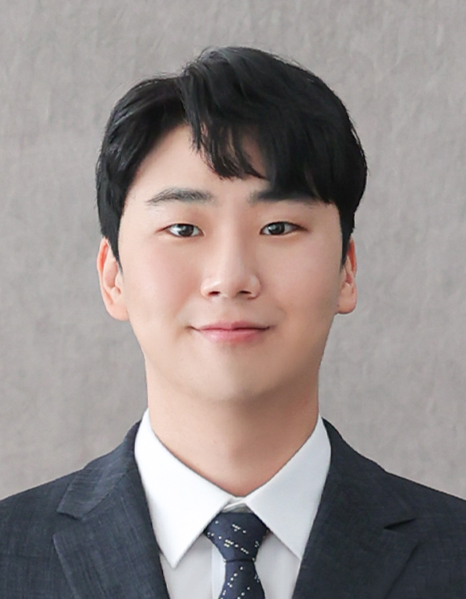}}]{\textit{Hyunsoo Kim}} (hskim@islab.snu.ac.kr) received the Ph.D. degree in Electrical and Computer Engineering with SNU, Seoul, South Korea, in 2025. 
Since 2025, he has been with Device Solution (DS) in Samsung Electronics, Suwon.
His research interests include deep learning-based wireless communications and vehicle-to-everything (V2X) communication.
\end{IEEEbiography}

\begin{IEEEbiography}[{\includegraphics[draft=false,width=1in,height=1.5in,clip,keepaspectratio]{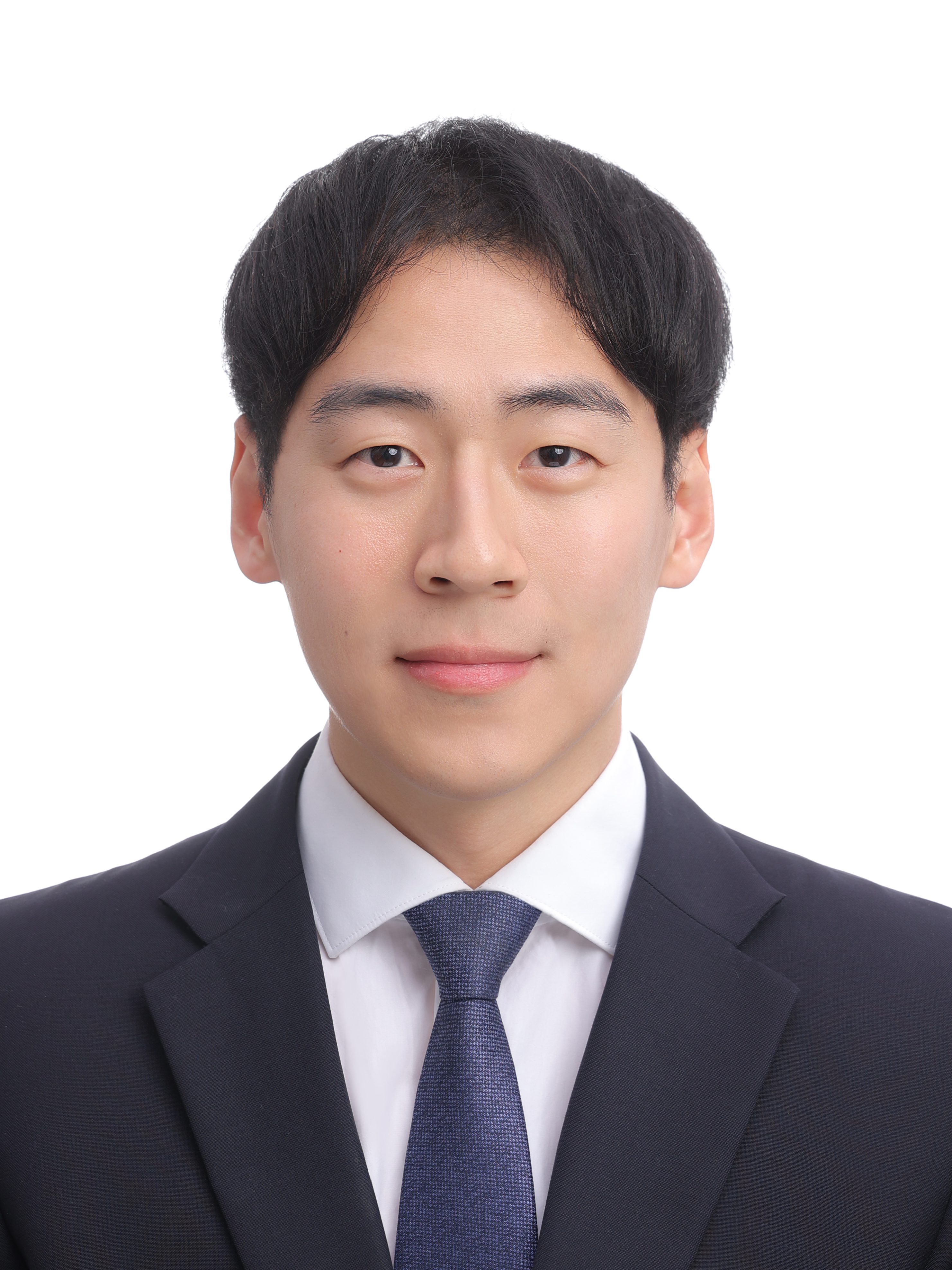}}]{\textit{Hyeonho Noh}} (hhnoh@hanbat.ac.kr) is an Assistant Professor with the Department of Information and Communication Engineering, Hanbat National University, Daejeon, South Korea. He received the Ph.D. degree in electrical engineering with the Pohang University of Science and Technology, Pohang, South Korea, in 2024. His research interests include integrated sensing and communication and signal processing with deep neural networks.
\end{IEEEbiography}

\begin{IEEEbiography}[{\includegraphics[draft=false,width=1in,height=1.5in,clip,keepaspectratio]{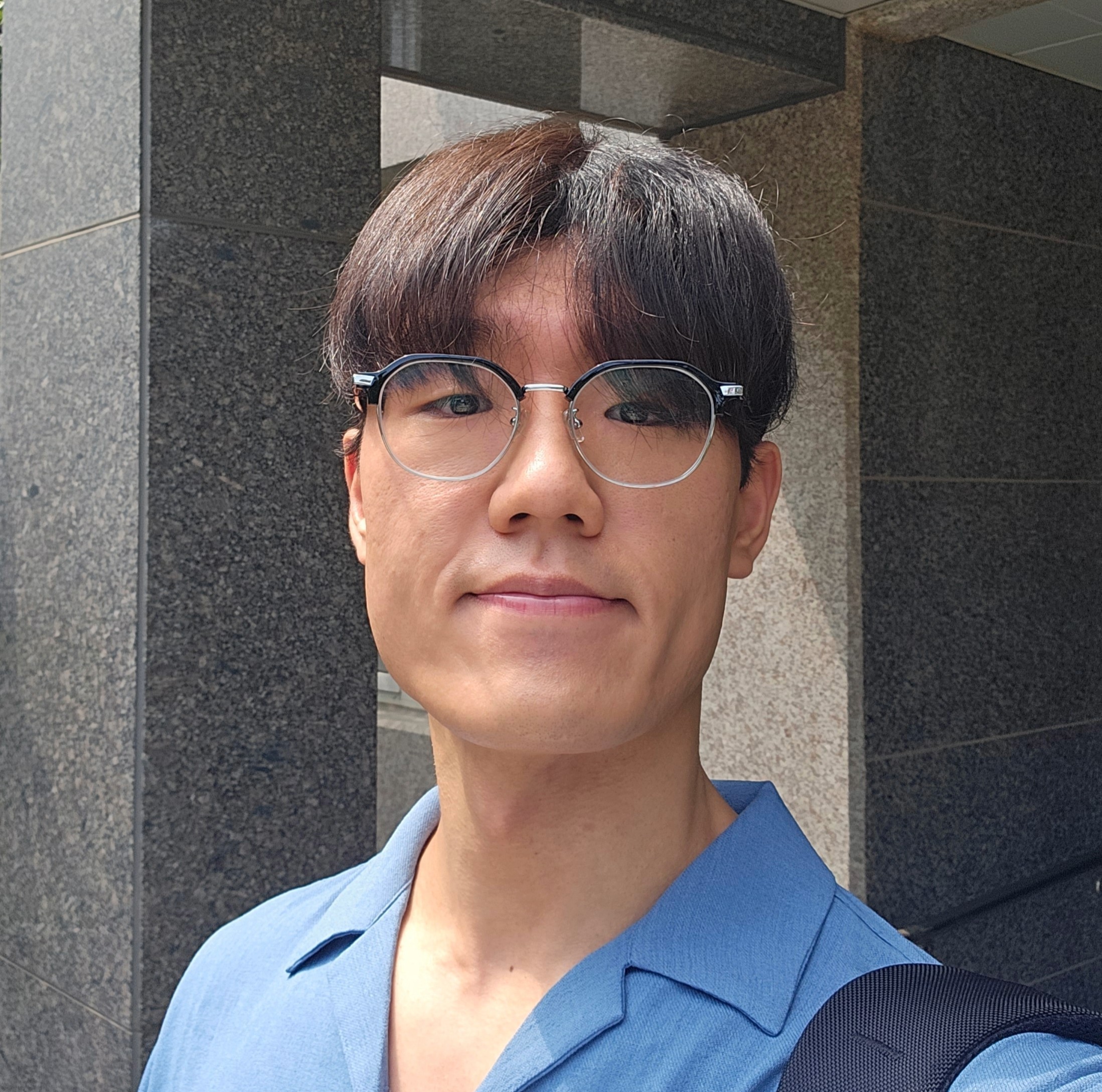}}]{\textit{Seungnyun Kim}} (snkim94@mit.edu) is a Postdoctoral Fellow with the Wireless Information and Network Sciences Laboratory, Massachusetts Institute of Technology, Cambridge, MA, USA. He received the Ph.D. degree in electrical and computer engineering from SNU, Seoul, South Korea, in 2023. His research interests include information theory, optimization methods, and machine learning for wireless communications.
\end{IEEEbiography}

\begin{IEEEbiography}[{\includegraphics[draft=false,width=1in,height=1.5in,clip,keepaspectratio]{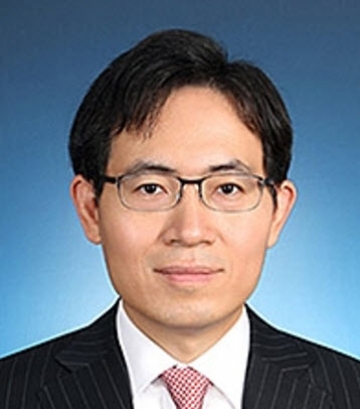}}]{\textit{Byonghyo Shim}} (bshim@islab.snu.ac.kr) is currently a Professor with the Department of Electrical and Computer Engineering, SNU, Seoul, South Korea. He is a Fellow of IEEE.
\end{IEEEbiography}

\end{document}